\documentclass[10pt,twocolumn,letterpaper]{article}

\usepackage{arxiv}
\newtoggle{usetimes}
\newtoggle{itabs}
\toggletrue{usetimes}  % Comment off this line to disable times font.
\iftoggle{usetimes}{\usepackage{times}}{}
\usepackage{xcolor}
\usepackage{amsmath,amssymb,amsfonts,dsfont,pifont,bm,bbm,mathrsfs,mathtools,nicefrac}
\usepackage{algorithm,algpseudocode,listings}
\usepackage{booktabs,multirow,adjustbox,diagbox,threeparttable}
\definecolor{citeblue}{HTML}{0071bc}
\usepackage[pagebackref=false,breaklinks=true,colorlinks=true,citecolor=citeblue,bookmarks=false]{hyperref}
\usepackage{cleveref}  % Should be loaded after 'hyperref', and works perfectly with 'subfigure'.
\usepackage{makecell}
\usepackage{stfloats}

\crefname{section}{Sec.}{Secs.}
\Crefname{section}{Section}{Sections}
\crefname{table}{Tab.}{Tabs.}
\Crefname{table}{Table}{Tables}
\crefname{figure}{Fig.}{Figs.}
\Crefname{figure}{Figure}{Figures}
\crefname{equation}{Eq.}{Eqs.}
\Crefname{equation}{Equation}{Equations}
\crefname{theorem}{Thm.}{Thms.}
\Crefname{theorem}{Theorem}{Theorems}
\crefname{algorithm}{Alg.}{Algs.}
\Crefname{algorithm}{Algorithm}{Algorithms}
\newcommand{\method}{SOONet\xspace}
\newcommand{\supp}{\textit{Appendix}}
\usepackage{listings}
\definecolor{codegreen}{rgb}{0,0.6,0}
\definecolor{codegray}{rgb}{0.5,0.5,0.5}
\definecolor{codepurple}{rgb}{0.58,0,0.82}
\definecolor{backcolour}{rgb}{1.0,1.0,1.0}
\lstdefinestyle{mystyle}{
    backgroundcolor=\color{backcolour},
    commentstyle=\color{codegreen},
    keywordstyle=\color{magenta},
    numberstyle=\tiny\color{codegray},
    stringstyle=\color{codepurple},
    basicstyle=\ttfamily\scriptsize,
    breakatwhitespace=false,
    breaklines=true,
    captionpos=b,
    keepspaces=true,
    numbers=left,
    numbersep=5pt,
    showspaces=false,
    showstringspaces=false,
    showtabs=false,
    tabsize=2
}
\begin{document}

\title{Scanning Only Once: An End-to-end Framework for Fast \\
Temporal Grounding in Long Videos}

\author{
    Yulin Pan$^{1}$ \quad
    Xiangteng He$^{2}$ \quad
    Biao Gong$^{1}$ \quad
    Yiliang Lv$^{1}$ \quad
    Yujun Shen$^{3}$ \quad
    Yuxin Peng$^{2}$ \quad
    Deli Zhao$^{1}$
    \\[5pt]
    $^1$Alibaba Group \quad
    $^2$Wangxuan Institute of Computer Technology, Peking University \quad
    $^3$Ant Group
    \\[5pt]
    {\tt\small \{yanwen.pyl, a.biao.gong, shenyujun0302, zhaodeli\}@gmail.com} \\
    {\tt\small \{hexiangteng, pengyuxin\}@pku.edu.cn} \quad
    {\tt\small yiliang.lyl@alibaba-inc.com} 
}

\maketitle

\begin{abstract}

Video temporal grounding aims to pinpoint a video segment that matches the query description.
Despite the recent advance in short-form videos (\textit{e.g.}, in minutes), temporal grounding in long videos (\textit{e.g.}, in hours) is still at its early stage.
To address this challenge, a common practice is to employ a sliding window, yet can be inefficient and inflexible due to the limited number of frames within the window.
In this work, we propose an end-to-end framework for fast temporal grounding, which is able to model an hours-long video with \textbf{one-time} network execution.
Our pipeline is formulated in a coarse-to-fine manner, where we first extract context knowledge from non-overlapped video clips (\textit{i.e.}, anchors), and then supplement the anchors that highly response to the query with detailed content knowledge.
Besides the remarkably high pipeline efficiency, another advantage of our approach is the capability of capturing long-range temporal correlation, thanks to modeling the entire video as a whole, and hence facilitates more accurate grounding.
Experimental results suggest that, on the long-form video datasets MAD and Ego4d, our method significantly outperforms state-of-the-arts, and achieves \textbf{14.6$\times$} / \textbf{102.8$\times$} higher efficiency respectively.
Project can be found at \url{https://github.com/afcedf/SOONet.git}.

\end{abstract}
\section{Introduction}\label{sec:intro}

\begin{figure}[t]
  \centering
  \includegraphics[width=\linewidth]{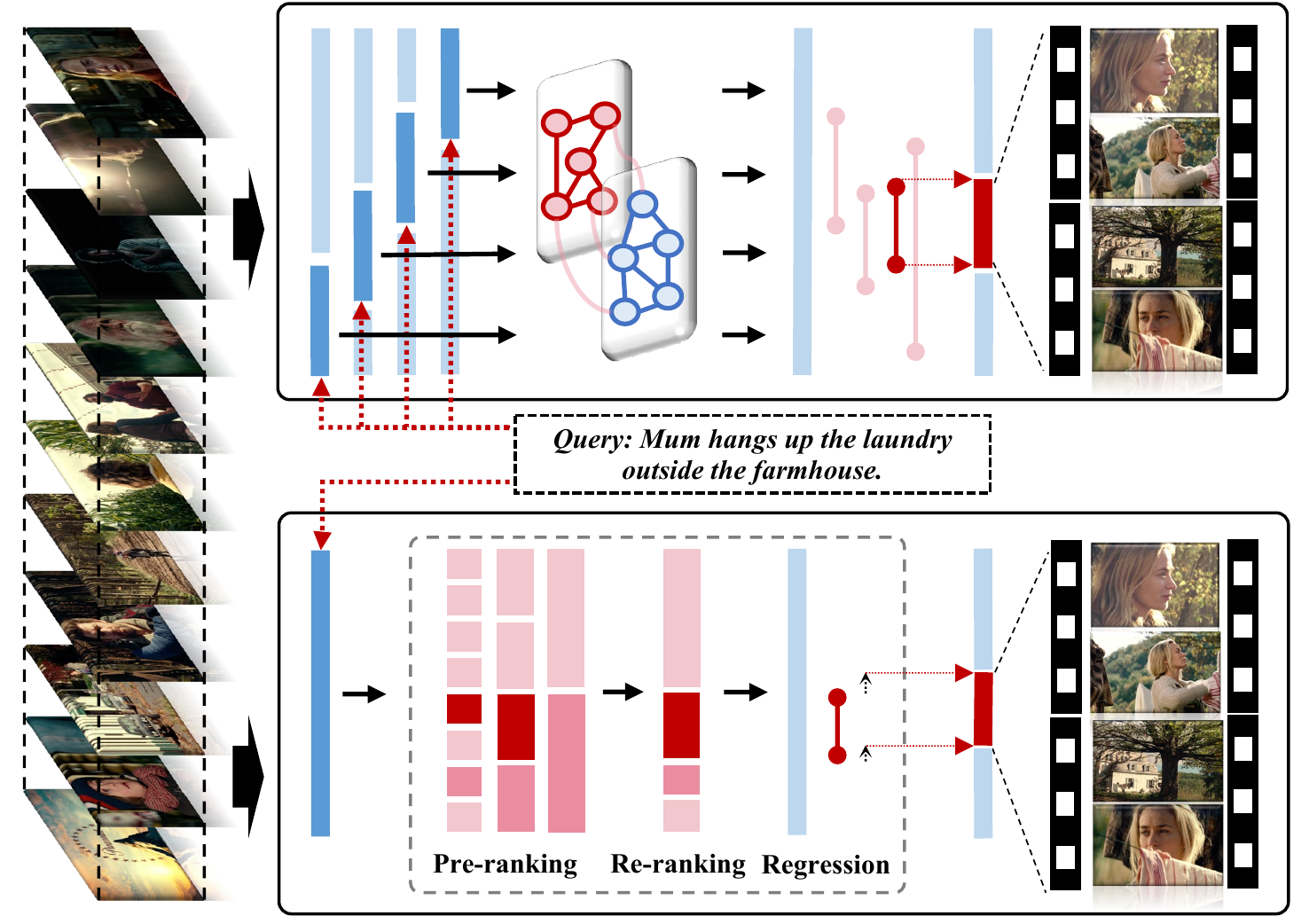}
  \caption{%
    \textbf{Pipeline comparison} between sliding window-based methods (top)~\cite{zhang2020learning,zeng2020dense,zhang2020span,soldan2021vlg} and our \method (bottom).
    It is noteworthy that the sliding window pipeline requires repeated inference on \textit{overlapped clips} and the final \textit{result aggregation}, while ours can deliver the result with \textit{one-time} network execution.
    Detailed discussion can be found in \cref{sec:efficiency}.
  }
  \label{fig:comparison}
  \vspace{-5mm}
\end{figure}

Video temporal grounding~\cite{zhang2020learning,zeng2020dense,gao2017tall,mun2020local,soldan2021vlg,zhang2020span,soldan2022mad,grauman2022ego4d}, which aims to localize a specific moment in the video corresponding to a natural language description, has found its applications in many real-world scenarios, such as video retrieval~\cite{lei2020tvr,wu2021hanet}, video highlight detection~\cite{yao2016highlight,xiong2019less}, and video question answering~\cite{huang2020location,yu2018joint}.
Despite the rapid advance in recent years, existing methods for temporal grounding usually target short-form videos (\textit{e.g.} in minutes) and characterize the input video with a small number of frames (\textit{e.g.}, 128)~\cite{zhang2020learning,zeng2020dense,zhang2020span,soldan2021vlg,zhang2019man,yuan2019semantic}.
When it comes to the case of long-form video temporal grounding (LVTG)~\cite{soldan2022mad,grauman2022ego4d}, however, temporally downsampling a video (\textit{e.g.}, in hours) to so few frames could cause severe information loss and further result in drastic performance degradation~\cite{grauman2022ego4d}.

A straightforward solution is to reorganize a long video to a sequence of short videos using a sliding window and perform temporal grounding within each window~\cite{soldan2022mad,grauman2022ego4d,hou2022cone}.
However, such a solution as shown in the top half of \cref{fig:comparison} has three main drawbacks.
(1) Inference inefficiency: The overlap between adjacent windows brings redundant computations. Besides, the large amounts of highly overlapped predictions cause post-processing (\textit{e.g.}, non-maximum suppression) time-consuming. It is noteworthy that by saying efficiency, we mean \textit{pipeline efficiency} instead of model efficiency, which considers the total execution time from data input to final result output, including data pre-processing, model forward running and post-processing.%
\footnote{The concrete explanation of each part can be found in \cref{sec:efficiency}.}
(2) Training insufficiency: The network with a sliding window can only scan video contents within a local time range at one time, yet ignore the long-range temporal correlation. 
(3) Prediction inflexibility: The prediction is restricted inside a single window, making it hard to generalize to segments with long duration.

In this work, we propose an anchor-based end-to-end framework, termed \textbf{\method}, which facilitates efficient and accurate LVTG by \textbf{S}canning a long-form video \textbf{O}nly \textbf{O}nce. 
As shown in the bottom half of \cref{fig:comparison}, \method follows a pipeline of \textit{pre-ranking}, \textit{re-ranking} and \textit{regression}, via leveraging both the inter-anchor context knowledge and the intra-anchor content knowledge. 

Specifically, we first produce non-overlapped anchor sequence via anchor partition layer, then three procedures are implemented to obtain final predictions: (1) Multi-scale context-based anchor features are acquired by modeling inter-anchor context knowledge via cascaded temporal swin transformer blocks~\cite{liu2021swin}. Meanwhile, a coarse anchor rank is obtained via sorting the context-based matching scores with respect to query. (2) Content-based anchor features and a content-enhanced anchor rank can be obtained by supplementing anchors with detailed intra-anchor content knowledge. We pick out the top-$m$ anchors that highly corresponds to query from each scale to form an anchor subset, then implement re-ranking within subset to reduce the computational complexity. (3) Boundary regression is adopted to achieve flexible predictions, leveraging both inter-anchor and intra-anchor knowledge.
To take full advantage of the abundant cross-modal semantic relationship in long videos, we sample one video with a batch of queries grounded in this video at one training step, then optimize the full-length anchor rank and query rank simultaneously with the help of proposed dual-form approximate rank loss, which achieves superior cross-modal alignment.
Extensive experiments are conducted on two long-form video datasets, \textit{i.e.}, MAD~\cite{soldan2022mad} and Ego4d~\cite{grauman2022ego4d}. Our method significantly outperforms state-of-the-arts, and achieves \textbf{14.6$\times$} / \textbf{102.8$\times$} higher pipeline efficiency, which verifies the effectiveness.

\section{Related Work}\label{sec:related-work}

\subsection{Short-form Video Temporal Grounding}
Existing methods mainly focus on short-form video temporal grounding and can be categorized into \textit{proposal-based} and \textit{proposal-free} methods. Methods in proposal-based category adopt a two-stage pipeline, which first generate proposal candidates by various proposal generation methods, such as sliding window and proposal generation network, then they rank these candidates and output the proposal with the highest matching score as final prediction. \cite{gao2017tall} propose CTRL, a pioneer work in video grounding. CTRL produces various-length proposal candidates via sliding window and uses the visual-textual fusion modules combined with three operators, \textit{i.e.}, add, multiply and fully-connected layer, to obtain multi-modal fused representation. MAN \cite{zhang2019man} and SCDM~\cite{yuan2019semantic} leverage multiple cascaded temporal convolution layers to generate proposal candidates hierarchically. TGN~\cite{chen2018temporally} temporally captures the evolving fine-grained frame-by-word interactions and uses pre-set anchors to produce multi-scale proposal candidates ending at each time step.
Subsequently \cite{zhang2019cross,wang2020temporally,qu2020fine,zhang2021multi} follow the anchor-based framework and propose various multi-modal reasoning strategies to achieve precise moment localization. In addition, 2D-TAN~\cite{zhang2020learning} enumerate all possible segments as proposal candidates and convert them into 2D feature map, then a temporal adjacent network is proposed to obtain multi-modal representation and encode the video context information. Following this, \cite{zheng2021progressive,wang2021structured,soldan2021vlg} design more complicated cross-modal reasoning strategies to learn the video-language semantic alignment from both coarse and fine-grained granularities. 

Methods in proposal-free category predict the start and end boundaries by computing the time pair directly, or output the confidence scores of being the start and end positions of target moment for each snippet in video. \cite{yuan2019find} propose ABLR, which performs cross-modal reasoning with a multi-modal co-attention interaction modules and outputs target moments by feeding the multi-modal features to regressor. Attention weight-based regression and attention feature-based regression are considered together to achieve precise boundary regression. Concurrently, DRN~\cite{zeng2020dense} considers the data imbalance issue and only uses the frame in ground-truth moment to mitigate the sparsity issue. LGI~\cite{mun2020local} aligns the video and language from phrase-level and propose a local-global interaction network that models the cross-modal relationship considering local and global context information simultaneously.

However, directly applying these methods on long-form videos results in drastic performance degradation, as temporally downsampling a long video to so few frames causes severe temporal information loss.

\subsection{Long-form Video Temporal Grounding}
Recently MAD~\cite{soldan2022mad} and Ego4d~\cite{grauman2022ego4d} pose the challenge of long-form video temporal grounding, and give some baselines that integrate sliding window and temporal downsampling into some short video-fit methods, such as 2D-TAN~\cite{zhang2020learning}, VLG-Net~\cite{soldan2021vlg} and VSLNet~\cite{zhang2020span}. However, all these methods achieve inferior performance, considering both accuracy and efficiency. 
Recently~\cite{hou2022cone} propose CONE, which pre-filters the candidate windows to address the inference inefficiency and learns the cross-modal alignment from proposal-level and frame-level. Nevertheless, it adopts sparse sampling strategy at training stage, which does not explore the potential of long-form video adequately. 
\begin{figure*}[t]
  \centering
  \includegraphics[width=0.88\linewidth]{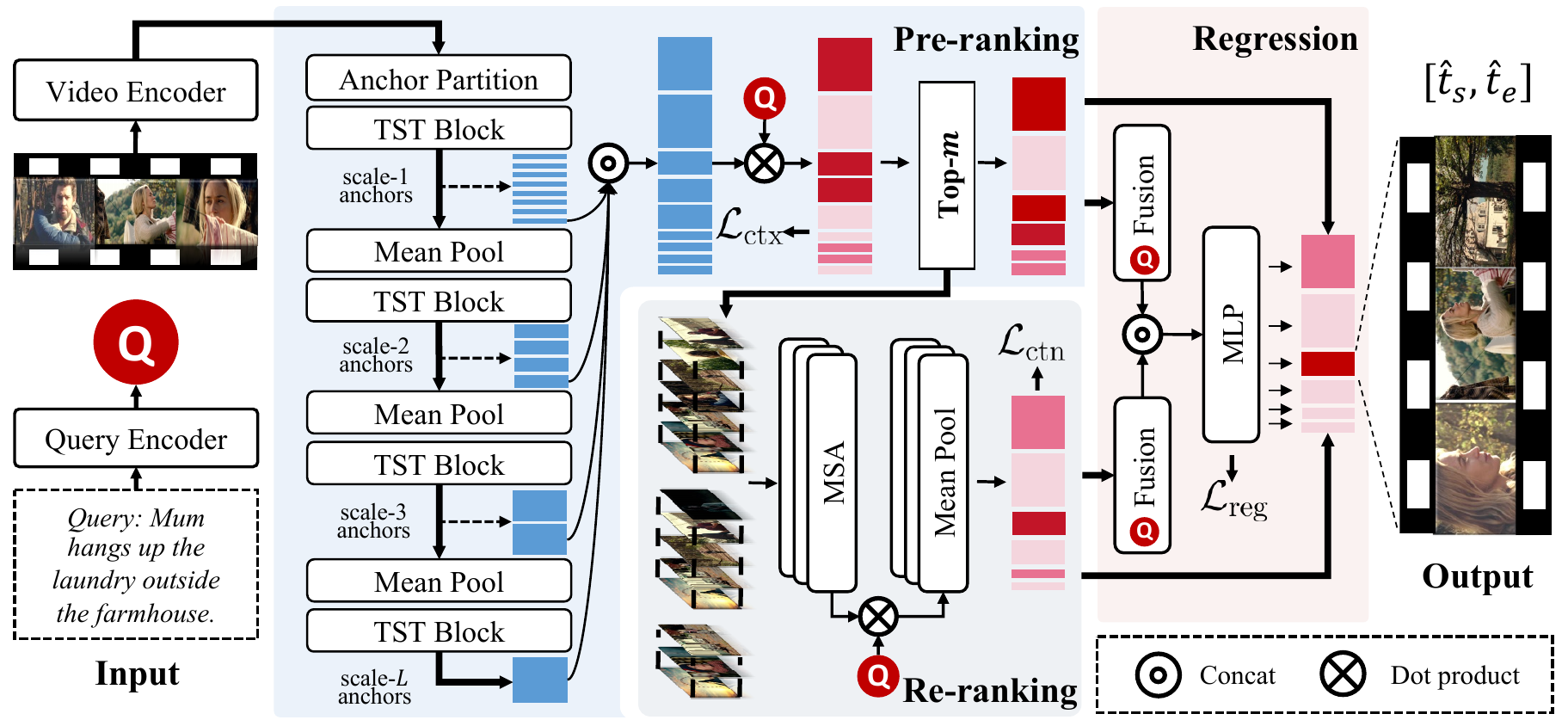}
  \caption{%
    \textbf{Overall architecture of our algorithm.} The whole framework consists of three modules: the \textit{pre-ranking module} aims to obtain coarse anchor rank by modeling inter-anchor context; the \textit{re-ranking module} aims to obtain content-enhanced anchor rank by supplementing anchors with detailed content; the \textit{regression module} aims to adjust anchor boundaries.
  }
  \label{fig:framework}
  \vspace{-5pt}
\end{figure*}

\section{Method}\label{sec:method}
This section presents a detailed introduction to our proposed framework.
As depicted in \cref{fig:framework}, our method takes a long-form video and a sentence query as input, and predicts the video moment that is semantically related to the query in an end-to-end manner. 
Specifically, our framework consists of three modules:
\textit{(1) Pre-ranking with Anchor Knowledge} aims to encode the inter-anchor context by employing cascaded temporal swin transformer blocks. 
Then, a coarse anchor rank is obtained by sorting the context-based matching scores concerning the query.
\textit{(2) Re-ranking with Frame Knowledge} is designed to model the intra-anchor content knowledge, and calculate the content-based matching scores concerning the query. The anchor candidates are re-ranked by summing the context-based and content-based matching scores.
\textit{(3) Boundary Regression} aims to adjust anchor boundaries, leveraging both inter-anchor context and intra-anchor content. 
Our method outputs the adjusted boundaries of the top-$n$ anchors as the final predictions.

\subsection{Feature Extractor}
Given an untrimmed video $V=\{f_{t}\}_{t=1}^{T}$ and a sentence query $Q=\{w_{m}\}_{m=1}^{M}$, where $T$ and $M$ represent the number of frames and words respectively, the LVTG task requires to localize the target moment $(\tau_{s}, \tau_{e})$ that corresponds to query. To achieve this, we adopt off-the-shelf pretrained models to extract visual features $\mathbf{V}=\{\mathbf{v}_{1}, \mathbf{v}_{2}, ..., \mathbf{v}_{N}\}\in\mathbb{R}^{N \times D}$ as well as textual features $\mathbf{Q}=\{\mathbf{q}_{\text{cls}}, \mathbf{q}_{1}, \mathbf{q}_{2}, ..., \mathbf{q}_{M}\}\in\mathbb{R}^{(M+1) \times D}$. $N,M$ represent the numbers of extracted frame features and word features respectively, and $D$ represents the feature dimension. 
The query feature $\mathbf{q}$ is extracted in different ways depended on the type of pretrained model.
For models pretrained with multi-modalities (\textit{e.g.}, CLIP~\cite{radford2021learning}), we take out the class token embedding $\mathbf{q}_{\text{cls}}$ as query feature. While for other models (\textit{e.g.}, BERT~\cite{devlin2018bert}), we pass the word embeddings through a trainable LSTM \cite{hochreiter1997long} layer to acquire the query feature.
We then feed the video features $\mathbf{V}$ and query feature $\mathbf{q}$ into our network for next process.

\subsection{Pre-ranking with Anchor Knowledge}
\noindent\textbf{Multi-scale anchor generation.}
Due to the computational complexity of global self-attention is quadratic to the sequence length,
the standard transformer is heavily computational on modeling full-length frame sequences of long-form video. 
To mitigate the computational burden, we first employs a single convolutional layer to produce non-overlapping base anchors from successive frames. The formulation is as follow:
\begin{equation}
    \mathbf{E}^{0} = \text{Conv1d}(\mathbf{V}),
\end{equation}
where $\mathbf{E}^{0} \in \mathbb{R}^{\frac{N}{C_{0}}\times D}$, and $C_{0}$ denotes the length of base anchor.
Then we adopt $L$ cascaded temporal swin transformer blocks with pooling layers to encode inter-anchor context knowledge and obtain multi-scale context-based anchor features $\mathbf{E} = [\mathbf{E}^{1}; \mathbf{E}^{2}; ...; \mathbf{E}^{L}]$, where $L$ represents the number of scales. Each anchor feature $\mathbf{e}_{i} \in \mathbf{E}$ corresponds to a unique clip proposal $(t_{s}^{i}, t_{e}^{i})$. For the anchors of $l$-th scale, the corresponding anchor length is
\begin{equation}
    C_{l} = C_{l-1}r_{l},
\end{equation}
where $r_{l}$ denotes the receptive field of $l$-th pooling layer.

\noindent\textbf{Temporal swin transformer block.}
We have incorporated the shifted window-based self-attention approach, as proposed in Swin Transformer~\cite{liu2021swin}, into 1-dimensional sequence encoding. 
This technique effectively implements self-attention in local windows, while also establishes connections between consecutive windows to bolster the modeling capabilities. 
In this way, the computational complexity is linearly scaling with the sequence length.
Specifically, each temporal swin transformer block consists of a local-window self-attention layer ($\text{W-MSA}$), a shifted-window self-attention layer ($\text{SW-MSA}$) and two multi-layer perceptrons ($\text{MLP}$), which can be formulated as:
\begin{equation}
\begin{split}
    & \hat{z}^{l} = \text{W-MSA}(\text{LN}(z^{l})) + z^{l}, \\
    & \Tilde{z}^{l} = \text{MLP}(\text{LN}(\hat{z}^{l})) + \hat{z}^{l}, \\
    & \Tilde{z}^{l+1} = \text{SW-MSA}(\text{LN}(\Tilde{z}^{l})) + \Tilde{z}^{l}, \\
    & z^{l+1} = \text{MLP}(\text{LN}(\Tilde{z}^{l+1})) + \Tilde{z}^{l+1},
\end{split}
\end{equation}
where $\text{LN}$ represents the LayerNorm~\cite{ba2016layer} operation. 

For each context-based anchor feature $\mathbf{e}_{i} \in \mathbf{E}$, the context-based matching score is obtained by computing the cosine similarity between anchor feature and query feature, then scaling it to [0, 1] via Sigmoid function:
\begin{equation}
    S_{\text{ctx}}^{i} = {\rm Sigmoid}(\frac{\mathbf{e}_{i} \cdot \mathbf{q}}{\Vert \mathbf{e}_{i} \Vert\Vert \mathbf{q} \Vert}), 1 \leq i \leq \sum\limits_{l=1}^{L} \frac{N}{C_{l}}.
\end{equation}
Finally a coarse anchor rank can be acquired by sorting $S_{\text{ctx}}$ in a descending order.

\subsection{Re-ranking with Frame Knowledge}
To mitigate the temporal information loss caused by the anchor partition and pooling operation in \textit{pre-ranking module}, the \textit{re-ranking module} models the detailed content inside anchors and re-rank anchor candidates. 
Given the coarse anchor rank, we first collect the indices of the top-$m$ anchors from each scale separately to set up an anchor subset, then for $i$-th anchor of $l$-th scale in this subset, we fetch the intra-anchor frame features $\mathbf{V}_{i}=\{\mathbf{v}_{i}^{k}\}_{k=1}^{C_{l}}$ and adopt standard multi-head self-attention module (\text{MSA}) to model the intra-anchor frame correlation:
\begin{equation}
    \hat{\mathbf{V}}_{i} = \text{MSA}(\text{LN}(\mathbf{V}_{i} + f_{\text{pos}}(\mathbf{V}_{i})) + \mathbf{V}_{i},
\end{equation}
where $f_{\text{pos}}$ is trainable positional embeddings used to inject positional information.
The content-based matching score of $i$-th anchor is obtained by first computing cosine similarity between each frame feature and query feature, then averaging frame-wise similarities and scaling it to $[0, 1]$ via Sigmoid function:
\begin{equation}
    S_{\text{ctn}}^{i} = {\rm Sigmoid}(\frac{1}{C_{l}}\sum\limits_{k=1}^{C_{l}}\frac{\hat{\mathbf{v}}_{i}^{k} \cdot \mathbf{q}}{\Vert \hat{\mathbf{v}}_{i}^{k} \Vert\Vert \mathbf{q} \Vert}), 1 \leq i \leq mL.
\end{equation}
We sum the context-based score and content-based score as the final matching score for re-ranking:
\begin{equation}
    S = \Tilde{S}_{\text{ctx}} + S_{\text{ctn}},
\end{equation}
where $\Tilde{S}_{\text{ctx}} \subsetneqq S_{\text{ctx}}$ is the context-based scores of subset.

\subsection{Boundary Regression}
To achieve flexible localization, the \textit{boundary regression module} is employed to adjust anchor boundaries inward or outward.
For $i$-th anchor of $l$-th scale in anchor subset, given context-based anchor feature $\mathbf{e}_{i}$ and content-based anchor feature $\hat{\mathbf{V}_{i}}$, we fuse them with query to obtain multi-modal fused feature, and pass it through a MLP header to predict the start and end bias:
\begin{equation}
\begin{split}
    & \mathbf{f}^{i} = [\mathbf{e}_{i} \odot \mathbf{q}; {\rm Att}(\hat{\mathbf{V}}_{i}) \odot \mathbf{q}], \\
    & (\delta_{s}^{i}, \delta_{e}^{i}) = {\rm MLP}(\mathbf{f}^{i}),
\end{split}
\end{equation}
where $\odot$ is element-wise multiplication. ${\rm Att}(\hat{\mathbf{V}}_{i})$ represents the self-attentive accumulation of $\hat{\mathbf{V}}^{i}$:
\begin{equation}
\begin{split}
    & \alpha_{i}^{k} = \mathbf{W}\hat{\mathbf{v}}_{i}^{k}, \\
    & \mathbf{a}_{i} = {\rm Softmax}([\alpha_{i}^{1}, \alpha_{i}^{2}, ..., \alpha_{i}^{C_{l}}]), \\
    & {\rm Att}(\hat{\mathbf{V}}_{i}) = \sum\limits_{k=1}^{C_{l}} \mathbf{a}_{i}^{k}\hat{\mathbf{v}}_{i}^{k},
\end{split}
\end{equation}
where $W \in \mathbb{R}^{1 \times D}$ is a learnable weight matrix. Then given the original anchor boundaries $(t_{s}^{i}, t_{e}^{i})$, we add the predicted start and end bias respectively to obtain adjusted boundaries:
\begin{equation}
\begin{split}
    & \hat{t}_{s}^{i} = t_{s}^{i} + \delta_{s}^{i} \times (t_{e}^{i} - t_{s}^{i}), \\
    & \hat{t}_{e}^{i} = t_{e}^{i} + \delta_{e}^{i} \times (t_{e}^{i} - t_{s}^{i}).
\end{split}
\end{equation}
Finally we output the adjusted boundaries $(\hat{t}_{s}, \hat{t}_{e})$ of the top-$n$ anchors as final predictions.

\subsection{Training}
Two loss terms are adopted to optimize the network: (1) Cross-modal alignment loss $\mathcal{L}_{\text{align}}$, and (2) Boundary regression loss $\mathcal{L}_{\text{reg}}$. The total loss is a weighted combination of the two loss terms:
\begin{equation}
    \mathcal{L}_{\text{total}} = \lambda_{1} \mathcal{L}_{\text{align}} + \lambda_{2} \mathcal{L}_{\text{reg}},
\end{equation}
where $\lambda_{1}$ and $\lambda_{2}$ are hyper-parameters used to control the contribution of $\mathcal{L}_{\text{align}}$ and $\mathcal{L}_{\text{reg}}$ respectively.

\subsubsection{Cross-modal Alignment Loss}
We define the cross-modal alignment loss as a combination of context-based alignment loss $\mathcal{L}_{\text{ctx}}$ and content-based alignment loss $\mathcal{L}_{\text{ctn}}$:
\begin{equation}
    \mathcal{L}_{\text{align}} = \mathcal{L}_{\text{ctx}} + \mathcal{L}_{\text{ctn}}.
\end{equation}
For $\mathcal{L}_{\text{ctx}}$ and $\mathcal{L}_{\text{ctn}}$, we propose a dual-form approximate rank loss that adopts two ApproxNDCG~\cite{qin2010general} loss terms to optimize the anchor rank and query rank simultaneously. We first revisit the ApproxNDCG loss and introduce the dual-form approximate rank loss, then give out formal definitions of $\mathcal{L}_{\text{ctx}}$ and $\mathcal{L}_{\text{ctn}}$.

\noindent\textbf{ApproxNDCG loss.}
Given large amounts of anchor candidates, we aim to obtain such an anchor rank: the anchor semantically related to query should be ranked in front of the unrelated ones. To achieve this goal, rather than point-wise or pair-wise rank losses which are commonly used in existing methods, we adopt the list-wise ApproxNDCG loss to optimize the anchor rank from the global perspective: 
\begin{equation}
    \mathcal{L}_{ar}(S, y) = 1 - Z_{m}^{-1} \sum\limits_{i=1}^{K} \frac{2^{y_{i}}-1}{\log (1+\hat{\pi}_{i})},
\end{equation}
where $S$ denotes the matching scores of anchor candidates, $K$ is the number of anchor candidates and $Z_{m}$ refers to the discounted cumulative gain of the best rank. $y_{i}$ represents the matching degree between the $i$-th anchor and query that equals to the temporal IoU of their bounding boxes:
\begin{equation}
    y_{i} = {\rm IoU}((t_{s}^{i}, t_{e}^{i}), (\tau_{s}, \tau_{e})).
\end{equation}
$\hat{\pi}_{i}$ is a differentiable approximation to the rank of $i$-th anchor:
\begin{equation}
    \hat{\pi}_{i} = 1+ \sum\limits_{u \neq i}\frac{\exp(-\alpha(S_{i}-S_{u}))}{1+\exp(-\alpha(S_{i}-S_{u}))},
\end{equation}
where $\alpha$ denotes a temperature parameter. For each anchor, the ApproxNDCG loss compares it with all other anchors to decide its rank, taking full advantage of the semantic relationship in long-form videos.

\noindent\textbf{Dual-form approximate rank loss.} Besides the anchor rank optimization, considering the unique characteristic of long-form video dataset, we introduce an \textit{``one video with batch queries''} data sampling strategy that samples one video with a batch of queries grounded in this long video at one training step, and employ another ApproxNDCG loss to optimize the query rank simultaneously: 
\begin{equation}
    \mathcal{L}_{dar}(S^{a}, S^{q}, y) = \mathcal{L}_{ar}(S^{a}, y) + \mathcal{L}_{ar}(S^{q}, y),
\end{equation}
where $S^{a}$ and $S^{q}$ denotes the matching scores of anchor candidates and query candidates, respectively.
Now, we define the context-based alignment loss $\mathcal{L}_{\text{ctx}}$ and content-based alignment loss $\mathcal{L}_{\text{ctn}}$ as :
\begin{equation}
\begin{split}
    & \mathcal{L}_{\text{ctx}} = \mathcal{L}_{dar}(S_{\text{ctx}}^{a}, S_{\text{ctx}}^{q}, y), \\
    & \mathcal{L}_{\text{ctn}} = \mathcal{L}_{dar}(S_{\text{ctn}}^{a}, S_{\text{ctn}}^{q}, y),
\end{split}
\end{equation}
where $S_{\text{ctx}}^{a}$ and $S_{\text{ctn}}^{a}$ represents the full-length context-based anchor matching scores and $mL$-length content-based anchor matching scores respectively. Likewise, $S_{\text{ctx}}^{q}$ and $S_{\text{ctn}}^{q}$ denotes the context-based and content-based query matching scores respectively.

\subsubsection{Boundary Regression Loss}
We define the boundary regression loss as follows:
\begin{equation}
    \mathcal{L}_{\text{reg}} = \frac{1}{mL} \sum\limits_{i=1}^{mL} \mathcal{L}_{iou}((\hat{t}_{s}^{i}, \hat{t}_{e}^{i})),
\end{equation}
where the $(\hat{t}_{s}^{i}, \hat{t}_{e}^{i})$ is the adjusted boundaries of $i$-th anchor. IoU loss~\cite{yu2016unitbox} is adopted to regress the start and end bias between anchor boundaries and groundtruth moment:
\begin{equation}
    \mathcal{L}_{iou}((\hat{t}_{s}^{i}, \hat{t}_{e}^{i})) = -\ln ({\rm IoU}((\hat{t}_{s}^{i}, \hat{t}_{e}^{i}), (\tau_{s}, \tau_{e})).
\end{equation}
\section{Experiments}\label{sec:exp}
% MAD的实验结果
\begin{table*}[tbp]
  \centering\small
  \caption{%
    Performance on the test set of MAD dataset. All of three baselines are sliding window-based methods.
  }
  \vspace{-3pt}
  \setlength{\tabcolsep}{0.8mm}{
    \begin{tabular}{lcccccrcccccrccccc}
    \toprule
    \multirow{2}{*}{Model} & \multicolumn{5}{c}{\textbf{IoU = 0.1}} &       & \multicolumn{5}{c}{\textbf{IoU = 0.3}} &       & \multicolumn{5}{c}{\textbf{IoU = 0.5}} \\
    \cmidrule{2-6}\cmidrule{8-12}\cmidrule{14-18}  & R@1 & R@5 & R@10 & R@50 & R@100 & & R@1 & R@5 & R@10 & R@50 & R@100 & & R@1 & R@5 & R@10 & R@50 & R@100 \\
    \midrule
    VLG-Net~\cite{soldan2021vlg} & 3.64  & 11.66 & 17.89 & 39.78 & 51.24 &  & 2.76  & 9.31  & 14.65 & 34.27 & 44.87 &  & 1.65  & 5.99  & 9.77  & 24.93 & 33.95 \\
    CLIP~\cite{radford2021learning}    & 6.57  & 15.05 & 20.26 & 37.92 & 47.73 &  & 3.13  & 9.85  & 14.13 & 28.71 & 36.98 &  & 1.39  & 5.44  & 8.38  & 18.80 & 24.99 \\
    CONE~\cite{hou2022cone}    & 8.90  & 20.51 & 27.20 & 43.36 & -     &  & 6.87  & 16.11 & 21.53 & 34.73 & -     &  & 4.10  & 9.59  & 12.82 & 20.56 & -     \\
    \midrule
    \textbf{\method(Ours)} & \textbf{11.26}  & \textbf{23.21} & \textbf{30.36} & \textbf{50.32} & \textbf{58.66} &  & \textbf{9.00}  & \textbf{19.64} & \textbf{26.00} & \textbf{44.78} & \textbf{53.18} &  & \textbf{5.32}  & \textbf{13.14}  & \textbf{17.84} & \textbf{32.59} & \textbf{39.62} \\
    \bottomrule
    \end{tabular}
  }
  \label{tab:mad}%
  \vspace{-3mm}
\end{table*}

\subsection{Datasets}
We conduct experiments on two long-form video datasets MAD~\cite{soldan2022mad} (\textit{avg. 110.8 min / video}) and Ego4d~\cite{grauman2022ego4d} (\textit{avg. 25.7 min / video}), in which videos are much longer than those in previous datasets, such as ActivityNet Captions~\cite{krishna2017dense} (\textit{avg. 2.0 min / video}) and Charades-STA~\cite{sigurdsson2016hollywood} (\textit{avg. 0.5 min / video}).

\vspace{2pt}
\noindent\textbf{MAD} is a large-scale benchmark for long-form video temporal grounding, which contains over 384K natural language queries that derived from high-quality audio description of mainstream movies and grounded in over 1.2K hours of videos with very low coverage (an average duration of 4.1s). The length of videos in MAD ranges from 47 minutes to 202 minutes, which are orders of magnitude longer than previous datasets.

\vspace{2pt}
\noindent\textbf{Ego4d} is an egocentric video dataset, containing 3670 hours of daily-life activity videos collected by 931 worldwide participants. The \textbf{Ego4d-NLQ} is the official subtask of Ego4d which is to retrieve the most relevant video moment from truncated video clips, given a natural language question that generated via filling pre-defined query templates. However, the average duration of video clips is only \textit{8.25 minutes}, which is too short to be used as LVTG evaluation benchmark. To verify the effectiveness of our method on long-form video grounding, we introduce a new evaluation setting and name it \textbf{Ego4d-Video-NLQ}, where we replace the truncated video clips with full-length video, therefore the average duration of videos reaches \textit{25.7 minutes}. We report the performance on the validation set of Ego4d, under both Ego4d-NLQ and Ego4d-Video-NLQ settings.

\subsection{Metrics}
Following \cite{soldan2022mad, hou2022cone}, we adopt the standard metric ``Recall@$n$, IoU=$m$'' (R@$n$-$m$) for evaluation. Specifically, it represents the percentage of testing samples that have at least one grounding prediction whose IoU with groundtruth is larger than $m$ among top-$n$ predictions.

\subsection{Implementation Details}
Following \cite{soldan2022mad}, we use CLIP \cite{radford2021learning} to extract visual features and textual features for MAD dataset. We set $C_{0}=10$, $L=4$ for multi-scale anchor generation. $\lambda_{1}, \lambda_{2}$ are set to 1 and 20 respectively. $m$ is set to 100 for filtering. The temperature $\alpha$ for $\mathcal{L}_{\text{ctx}}$ and $\mathcal{L}_{\text{ctn}}$ are both set to 0.01. We train the network for 100k steps with an initial learning rate of 0.001, and decay it by a factor of 10 after 40k steps. When training, we set batch size as 32 (1 video with 32 queries grounded in this video at one step) and use AdamW as the optimizer. The feature dimension $D$ is set to 512.

For Ego4d-NLQ and Ego4d-Video-NLQ, we use the pre-extracted SlowFast features \cite{feichtenhofer2019slowfast} and Bert features \cite{devlin2018bert} as the visual and textual features, following \cite{grauman2022ego4d}. We set $C_{0}=1$, $L=7$ on Ego4d-NLQ and $C_{0}=6$, $L=4$ on Ego4d-Video-NLQ. $\lambda_{1}, \lambda_{2}$ are set to 1 and 5 respectively. $m$ is set to 20. We train the network for 30k steps with an initial learning rate of 0.0001, and decay it by 10 after 15k steps. Other hyper-parameters are the same as in MAD. All experiments are implemented on one A100 GPU with 80GB memory. 

\subsection{Accuracy Comparison with SOTAs}
We first compare our model with several state-of-the-art methods. \cref{tab:mad} reports the performance results on long video dataset MAD (the average video duration is around \textbf{110.8 minutes}) with three methods: CLIP~\cite{radford2021learning}, VLG-Net~\cite{soldan2021vlg} and CONE~\cite{hou2022cone}. All of them are sliding window-based methods. From \cref{tab:mad} we can observe that our method outperforms all other methods, achieving \textbf{2.13\%} and \textbf{1.22\%} performance gains, in terms of R@1-0.3 and R@1-0.5 respectively. Thanks to modeling the entire video as a whole, our method can capture long-range temporal correlation, and learn cross-modal alignment with abundant context information, which facilitates more accurate grounding. We also conduct experiments on Ego4d dataset and summarize the results on \cref{tab:ego4d}. We first compare performance under Ego4d-NLQ setting with three methods: 2D-TAN~\cite{zhang2020learning}, VSLNet~\cite{zhang2020span} and CONE~\cite{hou2022cone}. \cref{tab:ego4d} suggests that our method achieves competitive performance on Ego4d-NLQ, even though it tests on short-form videos (the average video duration is \textbf{8.25 minutes}). We then test the performance on Ego4d-Video-NLQ, where the average video duration is \textbf{25.7 minutes}. We re-implement the 2D-TAN and VSLNet with the public code released by \cite{grauman2022ego4d}: it combines the 2D-TAN with sliding window to fit long-form video while adopts the downsampling strategy for VSLNet to reduce the sequence length to 128. From \cref{tab:ego4d} we observe our SOONet achieves \textbf{2.20\%} / \textbf{0.98\%} performance gains in terms of R@1-0.3 and R@1-0.5 respectively, which demonstrates the effectiveness of our method on long-form video temporal grounding. 
%

% Ego4d的实验结果
\begin{table}[!tbp]
  \centering\small
  \caption{%
    Performance on the val set of Ego4d dataset, under Ego4d-NLQ and Ego4d-Video-NLQ settings. Noted that 2D-TAN and CONE are sliding window-based methods while VSLNet is downsampling-based method.
  }
  \vspace{-3pt}
  \setlength{\tabcolsep}{9pt}{
    \begin{tabular}{lcccc}
    \toprule
    \multirow{2}{*}{Model} & \multicolumn{2}{c}{\textbf{IoU = 0.3}} & \multicolumn{2}{c}{\textbf{IoU = 0.5}} \\
          & R@1    & R@5    & R@1    & R@5 \\
    \midrule
    \multicolumn{5}{c}{Ego4d-NLQ (\textit{avg. 8.25 min / video})} \\
    \midrule
    2D-TAN~\cite{zhang2020learning}  & 5.04  & 12.89 & 2.02  & 5.88 \\
    VSLNet~\cite{zhang2020span}  & 5.45  & 10.74 & 3.12  & 6.63 \\
    CONE\footnotemark ~\cite{hou2022cone}   & 10.40  & 22.74 & 5.03  & 11.87 \\
    % 2D-TAN$^{*}$    & 3.34  & 8.41 & 1.73  & 4.50 \\
    % VSLNet$^{*}$   & 4.83  & 9.91 & 2.81  & 6.12 \\ 
    % \midrule
    \method(Ours) & 8.00      & 22.40    & 3.76     & 11.09  \\
    \midrule
    \multicolumn{5}{c}{Ego4d-Video-NLQ (\textit{avg. 25.7 min / video})} \\
    \midrule
    2D-TAN~\cite{zhang2020learning}   & 1.70  & 4.59 & 0.82  & 2.77 \\
    VSLNet~\cite{zhang2020span}   & 1.57  & 4.44 & 0.75  & 2.22 \\
    % \midrule
    \textbf{\method(Ours)} &  \textbf{3.90}      &  \textbf{10.71}     & \textbf{1.80}      & \textbf{5.09}     \\
    \bottomrule
    \end{tabular}
  }
  \vspace{-5mm}
  \label{tab:ego4d}
\end{table}
\footnotetext{As for CONE, its code has not been released,  so for fair comparison we only report its result on Ego4d-NLQ provided by its original paper.}

\begin{table*}[tbp]
  \centering\small
  \caption{%
    \textbf{Efficiency comparison} on MAD and Ego4d-Video-NLQ.
    The total time is a summation of time of three parts: pre-processing, model forward, and post-processing. 
    For fair comparison, we feed one video and one query to the system at each time, and report the total running time over the entire test set.
    Compared to sliding window-based methods, which require repeated inference on overlapped clips and the final result aggregation (\textit{i.e.}, post-processing), our one-time execution pipeline is far more efficient.
  }
  \vspace{-3pt}
  \setlength{\tabcolsep}{5pt}{
    \begin{tabular}{clcccc|cccc}
    \toprule
    \multirow{2}{*}{\makecell[c]{Dataset}} & \multirow{2}{*}{\makecell[c]{Method}} & \multirow{2}{*}{\makecell[c]{Method Type}} & \multirow{2}{*}{\makecell[c]{Trainable\\Parameters}} & \multirow{2}{*}{\makecell[c]{FLOPs}} & \multirow{2}{*}{\makecell[c]{GPU\\Memory}} & \multicolumn{4}{c}{Execution Time (second)} \\
    & & & & & & Pre & Model & Post & Total \\
    \hline
    \multirow{3}{*}{MAD}    & CLIP~\cite{radford2021learning}    & Slide Window & 0     & 0.2G      & 2.9G  & 630.9s & 15.7s & 6741.2s & 7387.8s \\
                            & VLG-Net~\cite{soldan2021vlg} & Slide Window & 5,330,435   & 1757.3G   & 20.0G & 3350.3s & 10659.0s & 15546.7s & 29556.0s \\
                            & \textbf{\method(Ours)} & End-to-end & 22,970,947  & 70.2G     & 2.4G  & 42.4s & 438.9s & 23.7s & \textbf{505.0s} \\
    \hline
    \multirow{3}{*}{\makecell[c]{Ego4d-\\Video-NLQ}} & 2D-TAN~\cite{zhang2020learning}  & Slide Window  & 86,773,761 & 6160.0G & 3.9G & 442.1s & 2625.3s & 1153.7s & 4225.2s \\
                                        & VSLNet~\cite{zhang2020span}   & Down-sampling     & 866,435 & 0.9G & 2.8G & 10.7s & 56.9s & 1.4s & 69.0s \\
                                        & \textbf{\method(Ours)}       & End-to-end      & 25,203,779 & 5.4G & 1.8G & 16.7s & 23.6s & 0.8s & \textbf{41.1s} \\

    \bottomrule
    \end{tabular}
    }
  \vspace{-3mm}
  \label{tab:efficiency}
\end{table*}
\subsection{Efficiency Comparison with SOTAs}\label{sec:efficiency}
To evaluate the efficiency of our method, we compare SOONet with 3 sliding window-based methods (\textit{i.e.}, CLIP, VLG-Net and 2D-TAN) and 1 downsampling-based method (\textit{i.e.}, VSLNet) on MAD and Ego4d-Video-NLQ. Recall that the code of CONE is not publicly available so we can't make a fair comparison with it. As mentioned in \cref{sec:intro}, the efficiency here means pipeline efficiency, which considers the execution time of three parts: (1) \textit{pre-processing} (denoted as Pre), which transfers raw data to the form of model input; (2) \textit{model forward} (denoted as Model), which refers to network calculation; (3) \textit{post-processing} (denoted as Post), which drops highly overlapped predictions and acquire the top-$n$ segments.
\cref{tab:efficiency} reports the number of parameters, FLOPs, GPU memory usage of models and gives a detailed breakdown of execution time. 
For FLOPs and GPU memory usage, we measure them using same samples as input because they change with the length of input video. From \cref{tab:efficiency} we observe the GPU memory usages of sliding window-based methods surpass our SOONet obviously, because batch inference on local windows is adopted to accelerate the model forward. 
For execution time, at each time we feed one video and one sentence to the system and report the total execution time of each part separately over the entire test set. 
From \cref{tab:efficiency} we observe that compared with sliding window-based methods, our SOONet makes huge improvement on pipeline efficiency, achieving \textbf{14.6$\times$} / \textbf{58.5$\times$} / \textbf{102.8$\times$} higher inference speed, compared with CLIP, VLG-Net and 2D-TAN respectively.
It is noteworthy that model FLOPs only affects the model forward time. Though CLIP contains only a matrix multiplication operation that needs few FLOPs, it suffers from both the slow pre-processing, which needs to split an entire video into lots of overlapped windows as well as gather window features, and the slow post-processing, which employs NMS (\textit{i.e.}, non-maximum suppression) to abandon large amounts of highly overlapped predictions. 
In addition to the slow pre-processing and post-processing, an another efficiency bottleneck of sliding window-based methods lies in the redundant computation on overlapped windows, which increases the model FLOPs greatly, causing the model forward part time-consuming. 
Compared with downsampling-based VSLNet, our SOONet achieves competitive inference speed whereas a far superior accuracy. Despite a bit more FLOPs, our network spends less time on model forward running than VSLNet. These results demonstrate the efficiency of our method.

\subsection{Ablation Studies}

\noindent\textbf{Effectiveness of Each Module.}
We conduct experiments on MAD to verify the effectiveness of each module employed in our framework: (1) Pre-ranking with Anchor Knowledge, (2) Re-ranking with Frame Knowledge, and (3) Boundary Regression. We report the ablation results on \cref{tab:ablation_module}, where PR, RR, BR represent the three modules respectively. 
\cref{tab:ablation_module} suggest that, equipped with Pre-ranking module only, our method achieves 9.41\% / 7.07\% / 4.10\% performance in terms of R@1-0.1, R@1-0.3, R@-0.5 respectively, which is a competitive result compared with state-of-the-arts. Benefit from the long-range context encoding and global-view rank learning, the Pre-ranking module explore the cross-modal semantic relationship in long videos adequately, thus facilitates accurate grounding. Upon this, integrating Re-ranking module achieves improvements of +0.76\%/+0.58\%/+0.33\%, because the detailed frame knowledge supplement fine-grained semantics, \textit{e.g.}, the scene and objects occurred in few frames, that generally perturbed by many irrelated frames. Integrating Boundary Regression module achieves improvements of +1.38\%/+1.45\%/+0.69\%, which benefits from the flexible adjustments. The combination of the three modules achieves improvements of +1.62\%/+1.76\%/+1.13\%, which demonstrates the complementary of proposed modules.

\begin{table}[tbp]
  \centering\small
  \caption{%
    Ablation study on various modules in \method. PR, RR, and BR denote Pre-ranking module, Re-ranking module, and Boundary Regression module, respectively
  }
  \vspace{-3pt}
  \setlength{\tabcolsep}{1.5mm}{
    \begin{tabular}{ccc|cccccc}
    \toprule 
    \multirow{2}{*}{\textbf{PR}} & \multirow{2}{*}{\textbf{RR}}& \multirow{2}{*}{\textbf{BR}} & \multicolumn{2}{c}{\textbf{IoU = 0.1}} & \multicolumn{2}{c}{\textbf{IoU = 0.3}} & \multicolumn{2}{c}{\textbf{IoU = 0.5}} \\
    & & & R@1 & R@5 & R@1 & R@5 & R@1 & R@5 \\
    \midrule
    \checkmark & & & 9.41  & 20.68 & 7.07  & 17.02  & 4.10  & 11.08 \\
    \checkmark & \checkmark & & 10.17  & 21.94 & 7.65  & 17.98  & 4.43  & 11.37 \\
    \checkmark & & \checkmark & 10.79  & 22.37 & 8.52  & 18.73  & 4.79  & 12.00 \\
    \checkmark & \checkmark & \checkmark & \textbf{11.03}  & \textbf{22.99} & \textbf{8.83}  & \textbf{19.48}  & \textbf{5.23}  & \textbf{13.18} \\
    \bottomrule
    \end{tabular}
  }
  \vspace{-5pt}
  \label{tab:ablation_module}
\end{table}
\begin{table}[!t]
  \centering\small
  \caption{%
    Ablation study on dual-form approximate rank loss.
  }
  \vspace{-3pt}
  \setlength{\tabcolsep}{2.4mm}{
    \begin{tabular}{l|cccccc}
    \toprule
    \multirow{2}{*}{Loss} & \multicolumn{2}{c}{\textbf{IoU = 0.1}} & \multicolumn{2}{c}{\textbf{IoU = 0.3}} & \multicolumn{2}{c}{\textbf{IoU = 0.5}} \\
    & R@1 & R@5 & R@1 & R@5 & R@1 & R@5  \\
    \midrule
    $\mathcal{L}_{bce}$  & 0.05  & 0.51 & 0.01 & 0.10 & 0.00 & 0.01 \\
    $\mathcal{L}_{nce}$  & 5.26  & 13.65 & 4.09 & 10.90 & 2.32 & 6.73 \\
    $\mathcal{L}_{ar}$   & 10.08  & 22.02 & 8.15 & 18.47 & 4.80 & 12.04 \\
    \midrule
    $\mathcal{L}_{dar}$  & \textbf{11.03}  & \textbf{22.99} & \textbf{8.83} & \textbf{19.48} & \textbf{5.23} & \textbf{13.18} \\
    \bottomrule
    \end{tabular}
  }
  \vspace{-8pt}
  \label{tab:ablation_loss}
\end{table}

\vspace{2pt}
\noindent\textbf{Impact of Dual-form Approximate Rank Loss.}
To make clear the contribution of the proposed dual-form approximate rank Loss $\mathcal{L}_{dar}$, we compare it with three loss functions: (1) Binary cross entropy loss $\mathcal{L}_{bce}$, which uses IoU as labels to optimize the query-anchor matching scores; (2) Noise contrastive estimation loss $\mathcal{L}_{nce}$, which optimizes a hidden space where positive pairs are assigned close and negative pairs are pushed away. We selects the anchor with highest IoU as positive samples and others as negative samples; (3) Single ApproxNDCG loss $\mathcal{L}_{ar}$, which optimizes the anchor rank only. The results are summarized in \cref{tab:ablation_loss}. $\mathcal{L}_{bce}$ achieves poor performance in all metrics, mainly caused by the extremely imbalance of positive (which has IoU $>$ 0) and negative (which has IoU $=$ 0) samples, even though we have enlarge the weight of positive samples. Besides, $\mathcal{L}_{ar}$ and $\mathcal{L}_{dar}$ both surpass $\mathcal{L}_{nce}$ by a large margin, because $\mathcal{L}_{nce}$ only tries to distinguish the anchor with highest IoU from large amounts of anchor candidates, while $\mathcal{L}_{ar}$ and $\mathcal{L}_{dar}$ implement the anchor rank optimization from the global perspective, which needs to consider the relationship between each anchor pair. Finally, $\mathcal{L}_{dar}$ outperforms $\mathcal{L}_{ar}$ by 0.68\% / 0.43\% in terms of R@1-0.3 and R@1-0.5, demonstrating the complementary of query rank optimization and anchor rank optimization.

\begin{figure}[t]
  \centering
  \includegraphics[width=\linewidth]{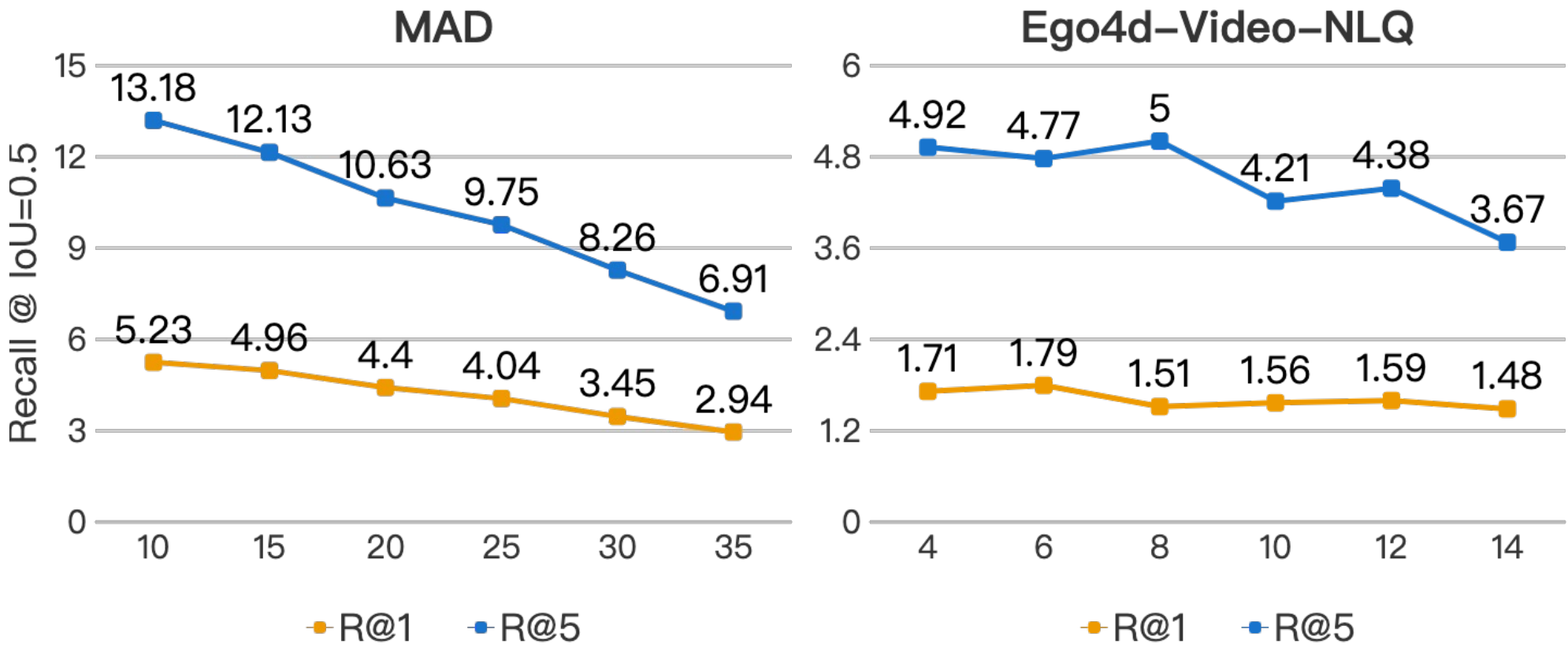}
  \caption{%
    Ablation study on the base anchor length, $C_{0}$. 
  }
  \label{fig:anchor}
  \vspace{-8pt}
\end{figure}
\begin{figure}[!t]
  \centering
  \includegraphics[width=\linewidth]{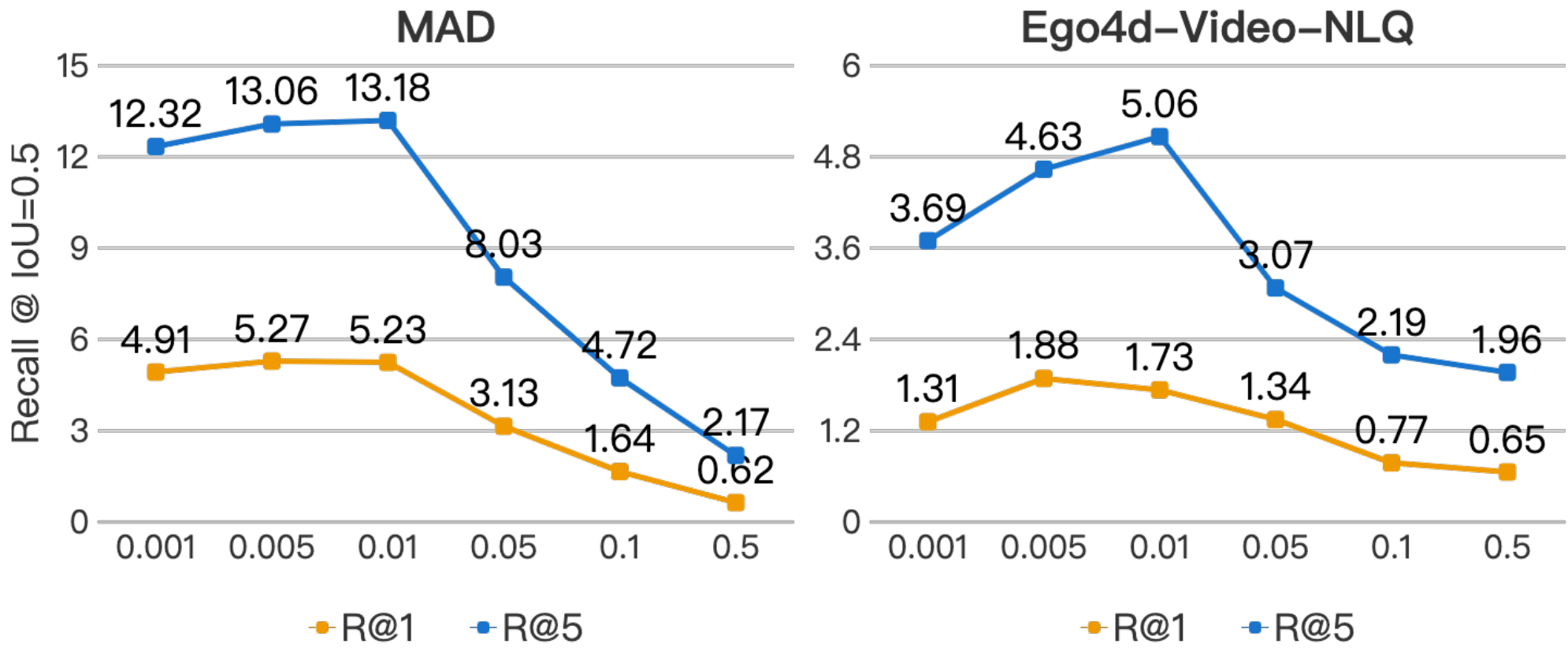}
  \caption{%
    Ablation study on the temperature value, $\alpha$.
  }
  \label{fig:tempe}
  \vspace{-5mm}
\end{figure}

\vspace{2pt}
\noindent\textbf{Base anchor length $C_{0}$.} We vary the value of $C_{0}$ to study the impact of anchor length and summarize the results in \cref{fig:anchor}. We observe that the performance decreases greatly on MAD as $C_{0}$ grows, while not changes obviously on Ego4d-Video-NLQ. 
This is because most of ground-truth moments on MAD last very short time, which makes the long anchor hard to align with query and regress the boundaries accurately. On the contrary, the length distribution of groundtruth in Ego4d-Video-NLQ is wider-ranging, which makes it insensitive to the anchor length.

\vspace{2pt}
\noindent\textbf{Temperature $\alpha$.} We vary the value of $\alpha$ in $\mathcal{L}_{dar}$ across from 0.001 to 0.5 to study the impact. The results are shown in \cref{fig:tempe}. From results we observe that the optimization is sensitive to the value of $\alpha$. The performance reaches the peak when $\alpha$ is in $[0.005, 0.01]$ and larger $\alpha$ leads to much worse performance.

\begin{figure}[t]
  \centering
  \includegraphics[width=\linewidth]{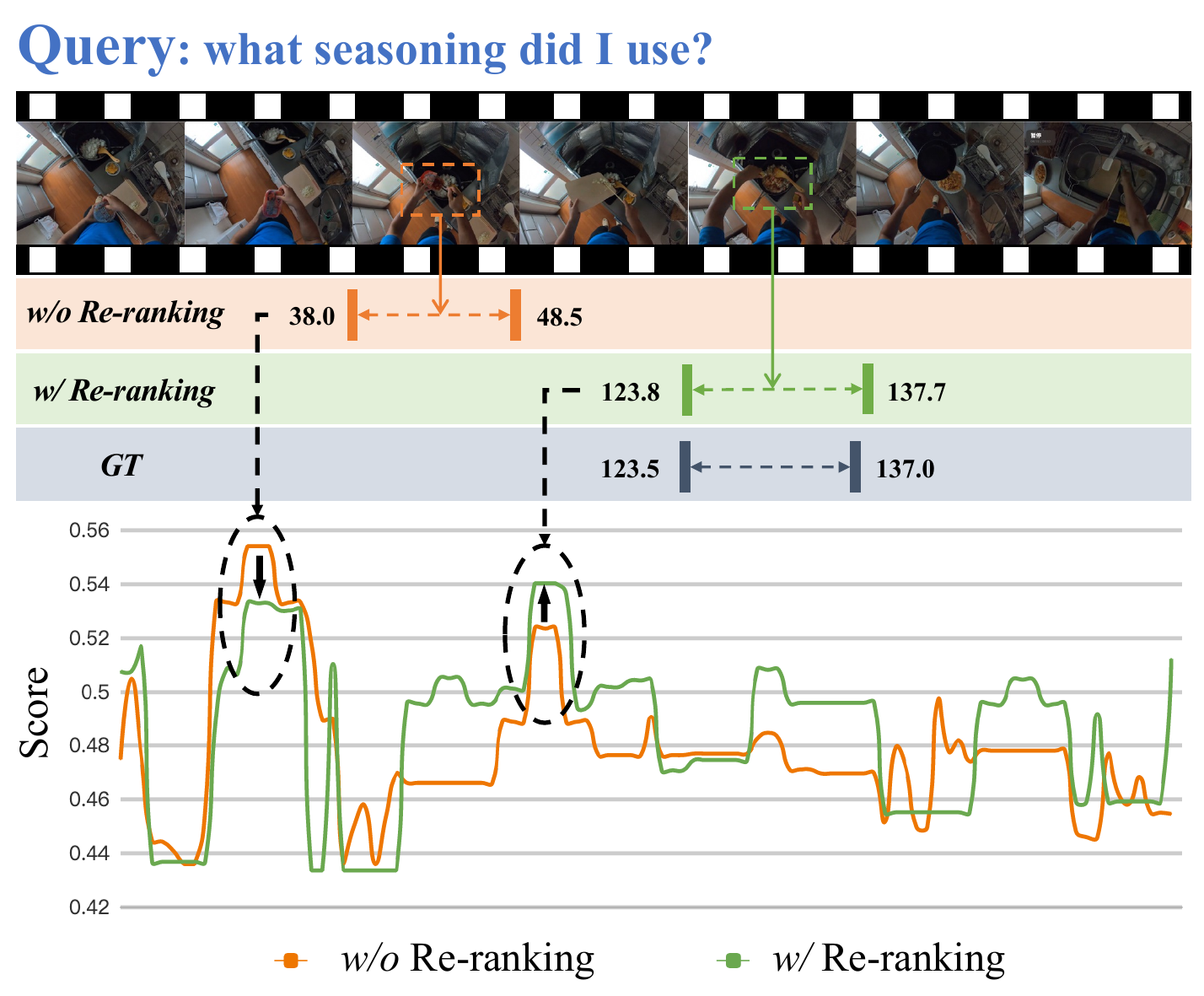}
  \caption{%
    Qualitative analysis on the re-ranking module with full-length anchor matching scores, where re-ranking helps localize the moment of interest more precisely.
  }
  \label{fig:visual}
  \vspace{-3mm}
\end{figure}

\subsection{Qualitative Analysis}
We provide qualitative results to illustrate the contributions of Pre-ranking and Re-ranking modules. \cref{fig:visual} displays the predictions of \method without Re-ranking (first line) and with Re-ranking (second line), as well as the corresponding groundtruth (third line). It suggests, equipped with only Pre-ranking module, our method achieves coarse localization (two humps showed on orange line) but loses some fine-grained details so that it can not distinguish food and seasoning. However, when combined with Re-ranking module, our method succeeds in recognizing the seasoning, and raises the confidence of the right moment as well as decreases the matching score of wrong moment. More qualitative results are provided in \supp.
\section{Conclusion}\label{sec:conclusion}
\renewcommand{\baselinestretch}{1.1}\normalsize
We propose an end-to-end framework, \method, for fast temporal grounding in long videos. It manages to model an hours-long video with one-time network execution, alleviating the inefficiency issue caused by the sliding window pipeline.
Besides, it integrates both inter-anchor context knowledge and intra-anchor content knowledge with carefully tailored network structure and training objectives, leading to accurate temporal boundary localization.
Extensive experiments on MAD and Ego4d datasets demonstrate the superiority of our \method regarding both accuracy and efficiency.

{\small
\bibliographystyle{abbrv}
\bibliography{ref}
}

\begin{figure*}[b]
  \centering
  \vspace{-8pt}
  \includegraphics[width=0.98\linewidth]{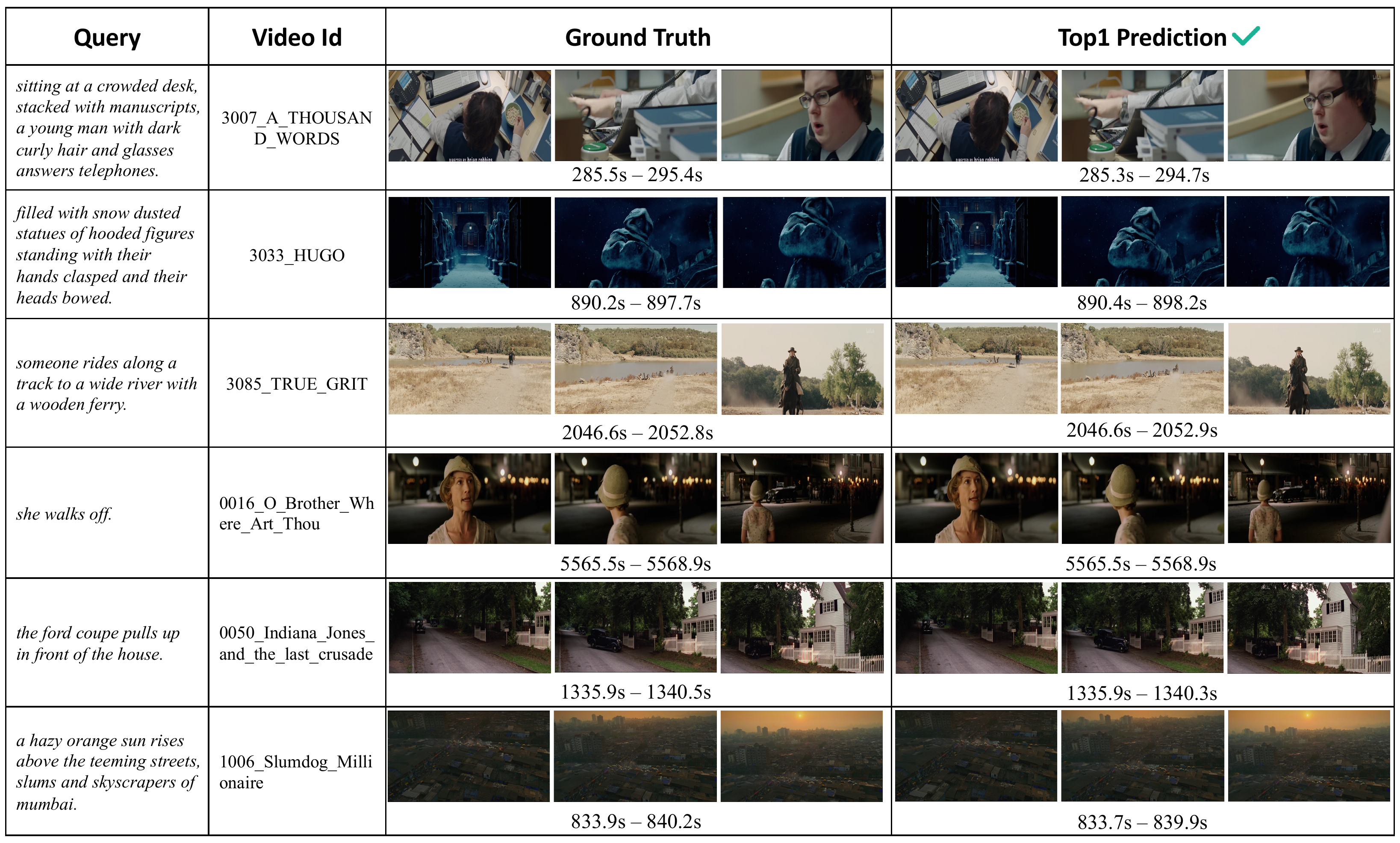}
  \vspace{-3mm}
  \caption{%
    \textbf{Visualization of some qualitative results on MAD.}
  }
  \label{fig:mad_success}
  \vspace{-5pt}
\end{figure*}

% \newpage
\appendix
\section*{Appendix}

\section{Qualitative Grounding Results}

\begin{figure*}[t]
  \centering
  \includegraphics[width=0.98\linewidth]{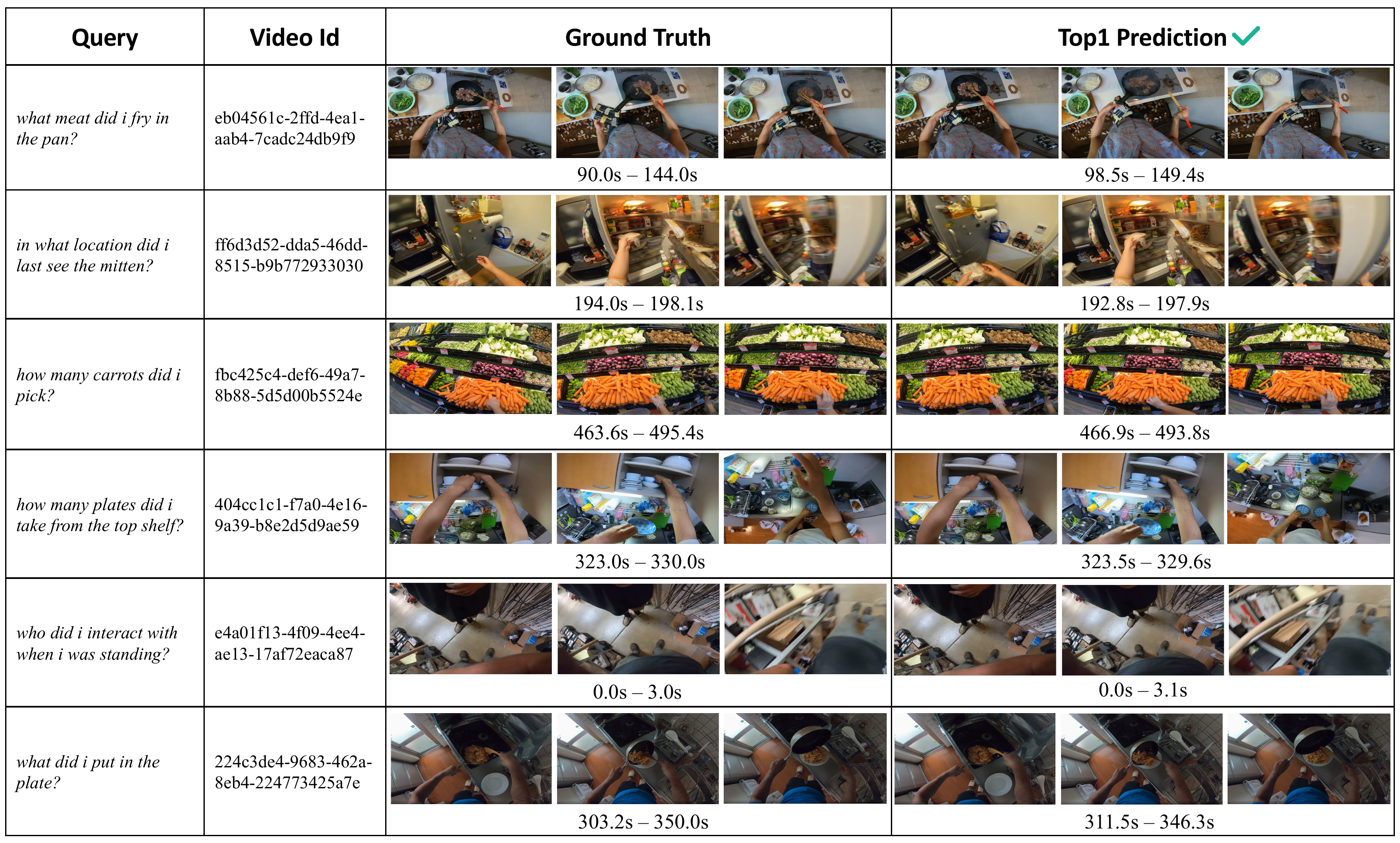}
  \vspace{-3mm}
  \caption{%
    \textbf{Visualization of some qualitative results on Ego4d-Video-NLQ.}
  }
  \label{fig:ego4d_success}
  \vspace{-2.5mm}
\end{figure*}

Recall that this work targets the temporal grounding in long videos. It proposes a tailored framework as well as training objectives, alleviating the inefficiency, insufficiency and inflexibility issues caused by the sliding window pipeline. 
We provide some qualitative results in \cref{fig:mad_success} and \cref{fig:ego4d_success} to illustrate the effectiveness of our method. They suggest, our method can achieve flexible boundary localization for various-length target segments.
Besides, thanks to the content-enhanced re-ranking, some inconspicuous objects which usually perturbed by other frames can be detected (\textit{e.g.}, the wooden ferry in the third sample of \cref{fig:mad_success}), hence facilitates accurate temporal localization.

\begin{figure*}[b]
  \centering
  \includegraphics[width=0.98\linewidth]{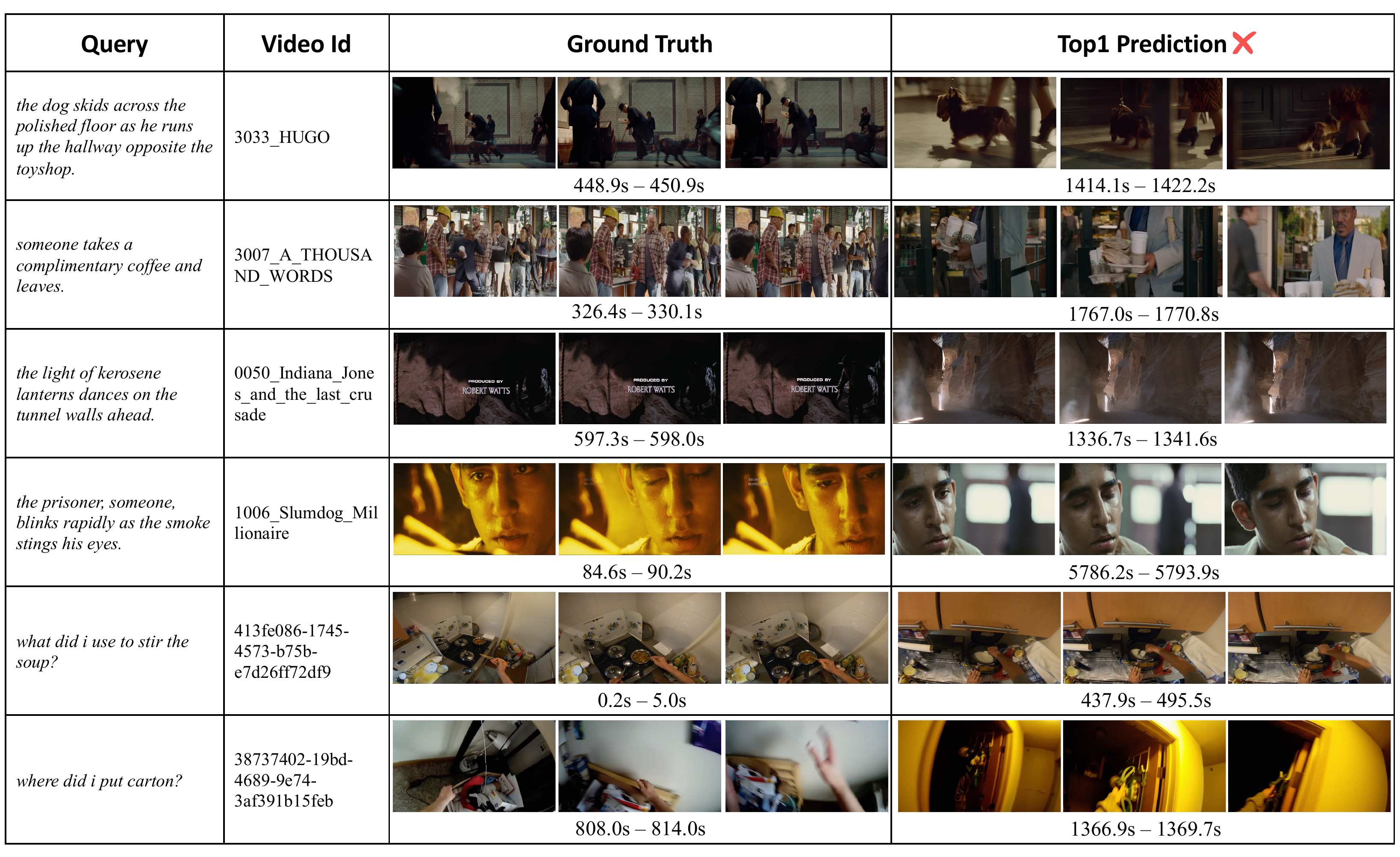}
  \vspace{-3mm}
  \caption{%
    \textbf{Visualization of some failure cases on MAD and Ego4d-Video-NLQ.}
  }
  \label{fig:failure}
\end{figure*}

Despite the effectiveness, due to the fact that the sentence feature is pre-extracted considering efficiency, some word-to-object alignment may get lost.
Here we provide some failure cases of our method in \cref{fig:failure}. We observe from top to bottom, our method misunderstands ``\textit{skids}'', ``\textit{a}'', ``\textit{kerosene lanterns}'', ``\textit{smoke}'', ``\textit{soup}'', ``\textit{carton}'' in turn.
They suggest that, the lack of explicit token-level semantic alignment learning leads to inadequate semantic analysis to some extent, which is left as our future work.

\end{document}